\documentclass[journal]{IEEEtran}

\usepackage{xcolor,soul,framed} 

\colorlet{shadecolor}{yellow}
\usepackage[pdftex]{graphicx}
\graphicspath{{../pdf/}{../jpeg/}}
\DeclareGraphicsExtensions{.pdf,.jpeg,.png}

\usepackage[cmex10]{amsmath}
\usepackage{amssymb}
\usepackage{array}
\usepackage{mdwmath}
\usepackage{mdwtab}
\usepackage{eqparbox}
\usepackage{url}
\usepackage{graphicx}
\usepackage{amsmath}
\usepackage{caption,subcaption}
\usepackage{hyperref}

\usepackage{algorithm} 
\usepackage{algpseudocode} 
\usepackage{algorithmicx}
\algrenewcommand\algorithmiccomment[1]{\hfill\(\triangleright\) #1}

\usepackage{array}
\usepackage{footnote}
\makesavenoteenv{tabular}
\usepackage{float}
\usepackage{subfloat}
\usepackage{multirow}

\begin{document}
\bstctlcite{IEEEexample:BSTcontrol}
    \title{Game-Theoretic Risk-Shaped Reinforcement Learning for Safe Autonomous Driving}
  \author{Dong Hu$^{1}$, Fenqing Hu$^{1}$, Lidong Yang$^{1}$, Chao Huang$^{1}$

  \thanks{Corresponding author: Chao Huang.}
  \thanks{$^{1}$Dong Hu, Fenqing Hu, Lidong Yang and Chao Huang are with the Department of Industrial and Systems Engineering, the Hong Kong Polytechnic University, Hong Kong (E-mail: dong24.hu@connect.polyu.hk; fengqing.hu@connect.polyu.hk; lidong.yang@polyu.edu.hk; hchao.huang@ polyu.edu.hk).}

}  

\markboth{   }{xx \MakeLowercase{\textit{et al.}}: xx}
\maketitle


\begin{abstract}
Ensuring safety in autonomous driving (AD) remains a significant challenge, especially in highly dynamic and complex traffic environments where diverse agents interact and unexpected hazards frequently emerge. Traditional reinforcement learning (RL) methods often struggle to balance safety, efficiency, and adaptability, as they primarily focus on reward maximization without explicitly modeling risk or safety constraints. To address these limitations, this study proposes a novel game-theoretic risk-shaped RL (GTR2L) framework for safe AD. GTR2L incorporates a multi-level game-theoretic world model that jointly predicts the interactive behaviors of surrounding vehicles and their associated risks, along with an adaptive rollout horizon that adjusts dynamically based on predictive uncertainty. Furthermore, an uncertainty-aware barrier mechanism enables flexible modulation of safety boundaries. A dedicated risk modeling approach is also proposed, explicitly capturing both epistemic and aleatoric uncertainty to guide constrained policy optimization and enhance decision-making in complex environments. Extensive evaluations across diverse and safety-critical traffic scenarios show that GTR2L significantly outperforms state-of-the-art baselines, including human drivers, in terms of success rate, collision and violation reduction, and driving efficiency. The code is available at \href{https://github.com/DanielHu197/GTR2L}{https://github.com/DanielHu197/GTR2L}.
\end{abstract}


\IEEEpeerreviewmaketitle

\section{Introduction}

\subsection{Background}

\IEEEPARstart{W}{ith} the rapid growth of intelligent transportation, autonomous driving (AD) is transforming the transportation sector. It promises improved road safety, reduced congestion, and support for smart cities \cite{11018512, hu2025toward}. However, real-world roads are complex and dynamic, with diverse traffic, frequent interactions, and sudden hazards. These challenges increase the difficulty of decision-making and demand greater safety and adaptability. Meeting these challenges still requires continued progress in AD \cite{cao2023continuous}.

In recent years, reinforcement learning (RL) has become a key approach for decision-making tasks due to its adaptability and self-learning capabilities \cite{chen2024end}. RL learns strategies through continuous interaction with the environment and shows strong potential in complex scenarios. It has already surpassed human performance in domains such as Go \cite{silver2016mastering}, online games \cite{vinyals2019grandmaster}, and drone racing \cite{kaufmann2023champion}. RL is thus widely used in AD for decision-making, motion planning, and navigation. It can generate either discrete decisions (e.g., lane changes) or continuous control commands (e.g., steering, acceleration) \cite{wu2024recent}. However, achieving efficient and robust long-term decision-making with RL while ensuring safety remains a major challenge \cite{yang2025human}. Standard RL focuses on maximizing expected rewards and cannot directly enforce hard constraints like collision avoidance or traffic rules, leading to safety risks in high-risk scenarios \cite{10924758}. As a result, safe RL has become a research focus to enhance the safety of AD.

\subsection{Safe Model-Free Reinforcement Learning}

Most research on safe RL focuses on model-free methods, as they do not require modeling environment dynamics \cite{hu2025autonomous}. A common approach is to solve constrained Markov decision processes (CMDPs) using the Lagrangian method, which converts safety constraints into unconstrained optimization problems \cite{ha2020learning}. Constrained policy optimization \cite{achiam2017constrained} is the first policy gradient method for CMDPs with monotonic improvement guarantees, but it is more computationally intensive and sensitive to approximation and sampling errors. Tessler et al.\cite{tessler2018reward} incorporated penalty terms into the reward function to address constraints, while Ding et al.\cite{ding2020natural} combined a primal-dual framework with natural policy gradients. Although CMDP-based methods are popular, they often fail to achieve zero constraint violations, primarily due to the sparsity of cost signals, which hinders effective learning as violations diminish. To address this, other model-free safe RL approaches utilize Lyapunov functions \cite{10795439}, control barrier functions (CBFs) \cite{10947350}, reachability analysis \cite{yu2022reachability,10919536}, game-theoretic formulations \cite{10637743}, or expert knowledge \cite{hu2024pre}. These methods typically rely on analytically or empirically designed safety certificates to constrain policy updates and ensure feasibility. However, such experiences or certificates are often manually designed \cite{donti2020enforcing}, and their effective application to complex tasks usually requires extensive trial-and-error.

\subsection{Safe Model-Based Reinforcement Learning}

Another class of safe RL methods learns an auxiliary world model to augment data, generate potential trajectories, or evaluate action safety prior to real-world execution \cite{sheng2024traffic}. The world model also facilitates the computation of safety certificates or policy constraints \cite{he2023fear}. These methods aim to improve sample efficiency while reducing risky exploration \cite{as2022constrained,jayant2022model}. For example, study \cite{as2022constrained} proposes a Bayesian world model to generate virtual trajectories and estimate optimistic and pessimistic bounds for objectives and constraints, respectively, using an augmented Lagrangian method to solve the constrained optimization. The Feariosity model \cite{11027413} incorporates curiosity and fear within a world model to guide safe and efficient exploration via adaptive policy constraints and threat-aware experience replay.

Model-based CBF methods rely on system dynamics models or expert safety knowledge. These methods introduce theoretical constraints to ensure the safety of the control process. For example, Robey et al. \cite{hunt2021verifiably} learn CBFs from expert trajectories in known nonlinear control-affine systems to constrain control inputs and keep the system within safe boundaries. In study \cite{emam2022safe}, the combination of RL and robust CBF improves the safety of systems under disturbances. In study \cite{cohen2023safe}, a safe exploration framework is built using Lyapunov-like CBFs to ensure the stability of the learning process. Although these methods can theoretically guarantee safety, they rely heavily on accurate system models. In complex scenarios, it is difficult to design such models, and they are easily affected by model errors, which limits their practical application.

\subsection{Safe Autonomous Driving}

Improving the safety of AD has always been a core aspiration of researchers. In recent years, many safe RL methods have been proposed. Safety checkers, as passive safety strategies, enhance safety through external mechanisms. For example, safety checkers based on kinematic models \cite{liu2022autonomous}, finite state machines \cite{hwang2022autonomous}, and safe action sets \cite{gu2023safe} have proven effective in certain scenarios. However, these strategies are highly based on expert knowledge and generally achieve safety at the cost of optimal performance.

Active safety strategies enable RL systems evaluate their reliability and trigger safety checks when needed. Hoel et al.~\cite{hoel2023ensemble} proposed policy aggregation to assess decision reliability and seek help if confidence is low. Cao et al.~\cite{cao2023continuous} proposed confidence-aware RL, letting the agent follow its policy only if it outperforms the rule-based one. Model-based RL is also gaining attention; for example, Kamran et al.~\cite{kamran2021minimizing} used a world model to simulate and select safer actions. As risk awareness becomes more important, traditional reward-based risk modeling often misses different risk types~\cite{10844066}. Li et al.~\cite{li2022decision} introduced a risk-probability-model-based RL approach for reducing driving risk, though it is limited to discrete actions. Huang et al.~\cite{huang2024safety} proposed a risk-aware shared-control RL approach that leverages dynamic safety checks to enhance both safety and robustness.

Some strategies improve RL safety by adding extra constraints. Value-based methods~\cite{wang2023autonomous} incorporate approximate safe action objectives for end-to-end control. For safety metrics, Li et al.~\cite{li2022decision} added risk terms to value functions for safer lane changes, while Yoo et al.~\cite{yoo2022gin} used predicted interactions as safety constraints in soft actor-critic (SAC). He et al.~\cite{he2022robust} introduced adversarial RL objectives for greater robustness, and another study~\cite{he2023fear} proposed a fear-inspired RL framework to encourage defensive driving. While these constraint-based methods show promise, their effectiveness is often scenario-specific, and applying them to more complex AD tasks still requires improving adaptability and robustness.

\begin{figure*}
    \begin{center}
    \includegraphics[width=0.9\linewidth]{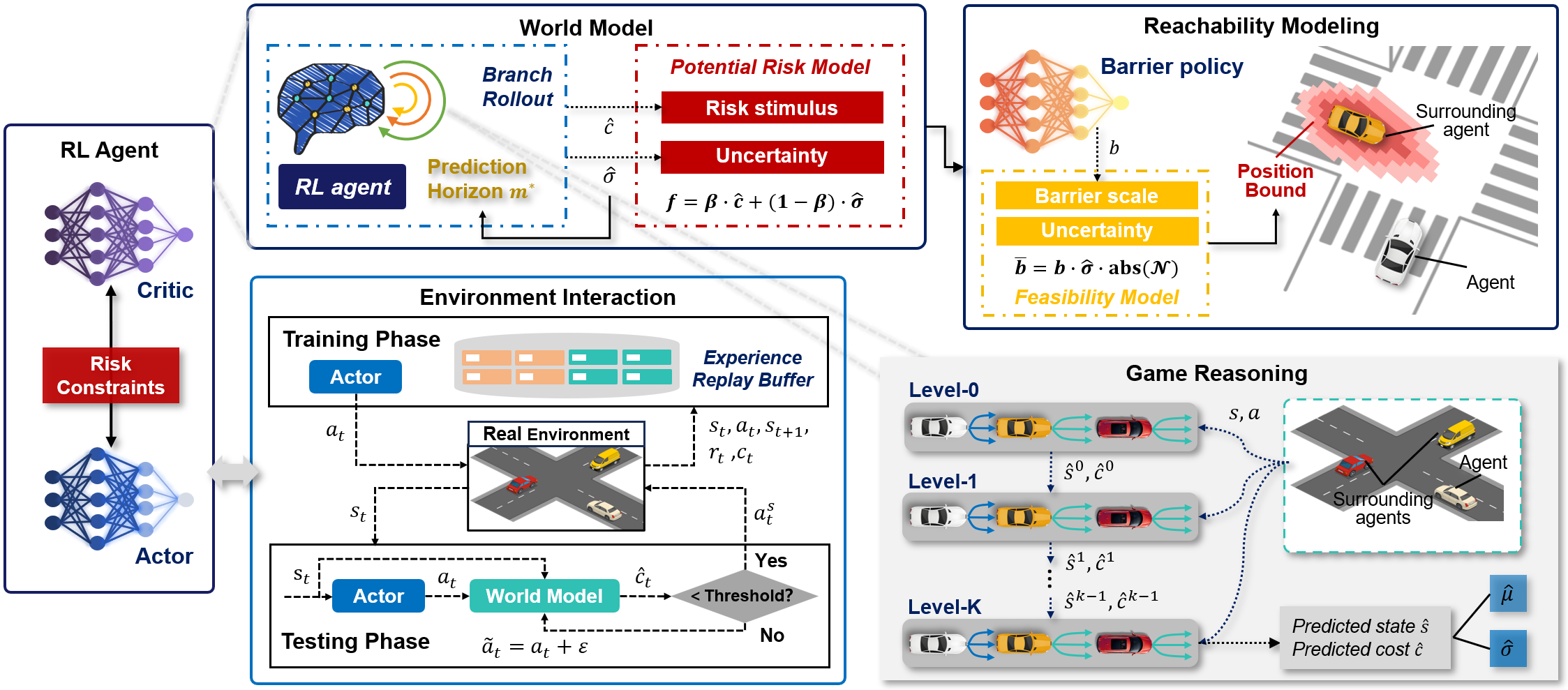}
    \caption{The framework of game-theoretic risk-shaped RL for safe autonomous driving, which leverages a game-theoretic adaptive-horizon world model, potential risk modeling, reachability (barrier) modeling, and risk-constrained RL to enhance safety and robustness in dynamic traffic environments.} \label{framework}
 \end{center}                                       
\end{figure*}

\subsection{Motivation and Contribution}

Existing methods face several key limitations. Current world models do not account for interactions with surrounding vehicles when making predictions, and typically use fixed prediction horizons, which limits adaptability to dynamic environments and complex situations. Barrier functions and risk-sensitive RL often rely on static constraints, reducing flexibility and robustness in the face of real-world uncertainties. Moreover, constraint functions in constrained RL are often coarse and fail to capture the uncertainties inherent in dynamic environments, leading to a suboptimal balance between safety and performance.

To tackle the above challenges, we propose \textbf{g}ame-\textbf{t}heoretic \textbf{r}isk-shaped \textbf{r}einforcement learning (GTR2L), a safe RL framework for AD. GTR2L combines a game-theoretic world model, uncertainty-aware reachability modeling, and long-term risk constraints with multi-level reasoning and adaptive horizon adjustment. The main contributions of this work are as follows:
\begin{itemize}
    \item A game-theoretic world model is proposed that integrates multi-level reasoning and adaptive horizon adjustment, enabling efficient modeling and inference in dynamic multi-agent environments;
    \item A dynamic barrier-based reachability modeling approach is developed, providing agents with flexible and adaptive safety driving boundaries in complex scenarios;
    \item A risk constraint mechanism is introduced, incorporating long-term risk feedback into policy optimization to promote proactive risk avoidance and improve robustness;
    \item Extensive experiments in diverse traffic scenarios show that the proposed method significantly enhances safety and traffic efficiency, outperforming mainstream approaches.
\end{itemize}

The remainder of this paper is organized as follows: Section~\ref{Sec_Preliminaries} provides the preliminaries. Section~\ref{Sec_Methodology} introduces the proposed framework and key techniques. Section~\ref{Sec_Experimental} details the experimental setup and evaluation metrics. Section~\ref{Sec_Results} presents results, comparisons, and discussion. Section~\ref{Sec_Conclusion} concludes and outlines future work.

\section {Preliminaries}\label{Sec_Preliminaries}
\subsection{RL-based Autonomous Driving}

In AD, driving tasks are often modeled as sequential decision-making problems, where the agent interacts with a dynamic environment to learn optimal and safe behaviors. This is typically formulated as a Markov decision process (MDP) with a state space $\mathcal{S}$ and an action space $\mathcal{A}$. The state space $\mathcal{S}$ includes ego-vehicle information, nearby vehicle states, and road context such as traffic light status and navigation goals. The action space $\mathcal{A}$ consists of actions like acceleration, deceleration, and lane changes. RL methods are used to learn policies that map states to actions, enabling autonomous vehicles to make safe and efficient decisions in complex traffic scenarios.

\subsection {Constrained Markov Decision Process}

A CMDP extends a standard MDP by introducing a cost function to capture safety or risk. Formally, a CMDP is defined by $(\mathcal{S}, \mathcal{A}, p, r, c, \gamma)$, where $p$ is the transition probability, $r$ is the reward function, $c$ is the cost function, and $\gamma$ is the discount factor.

The objective is to maximize the expected cumulative reward while keeping the expected cumulative cost below a given threshold. The CMDP optimization problem can be written as:
\begin{equation}
\max_{\pi} \; \mathbb{E} \left[ \sum_{t=0}^{\infty} \gamma^t r(s_t, a_t) \right],  \quad
 \text{s.t.} \quad \mathbb{E} \left[ \sum_{t=0}^{\infty} \gamma^t c(s_t, a_t) \right] \leq C,
\end{equation}
where $C$ is a predefined cost threshold.

\section {Methodology}\label{Sec_Methodology}

\subsection{Overview}

Fig.~\ref{framework} presents the overall architecture of the proposed safe AD framework, which tightly integrates game-theoretic world modeling, reachability modeling, and potential risk-based policy constraints. 

\textbf{World Model:} The world model uses multi-level game-theoretic reasoning to predict interactive behaviors and future states, outputting predictions and associated uncertainty. By jointly modeling risk and uncertainty, it enables the agent to better anticipate safety-critical situations, with the prediction horizon adaptively adjusted according to uncertainty for flexible planning.

\textbf{Reachability Modeling:} The reachability module employs a barrier policy to define the agent’s feasible region relative to surrounding vehicles, aiming to minimize potential risks. The safety boundary is dynamically adjusted according to the predictive uncertainty from the world model, enabling the agent to expand safety margins under high uncertainty and tighten them when confidence increases.

\textbf{RL Constraint:} The update process of RL agent is guided by potential risk constraints from the world model and adaptive reachable boundaries from the barrier model. This supports risk-aware policy optimization, ensuring the agent pursues long-term rewards while consistently maintaining safety under changing conditions and uncertainties.

\textbf{Interaction:} During testing, the matured world model provides real-time risk estimates for candidate actions as the agent executes its policy. If a candidate action is predicted to exceed the risk threshold, the agent adaptively modifies its decision or replans to maintain safety. This closed-loop process enables proactive risk avoidance and continuous adaptation to new or uncertain scenarios.

\subsection{Game-Theoretic World Model}

The world model provides an internal simulation of environment dynamics, enabling the agent to anticipate future states and assess risk under uncertainty. To this end, we adopt an ensemble of diagonal Gaussian models capable of capturing both aleatoric uncertainty and epistemic uncertainty~\cite{chua2018deep}. Formally, the ensemble is defined as:
\begin{equation}
\{\mathcal{M}_{\phi_n}\}_{n=1}^N, \quad \mathcal{M}_{\phi_n}(s',c|s,a) = \mathcal{N}(\mu_{\phi_n}(s,a), \sigma^2_{\phi_n}(s,a)),
\end{equation}
\noindent where $s'$ is the predicted next state, $N$ is the ensemble size, and $\mathcal{N}$ denotes a Gaussian distribution with mean $\mu_{\phi_n}(s,a)$ and variance $\sigma^2_{\phi_n}(s,a)$. Distinct from conventional reward-predictive models, our framework is designed for cost prediction, aligning with the needs of safety-critical applications.

The base model (Level-0) is trained by minimizing the negative log-likelihood of observed transitions:
\begin{equation}\label{update_WM}
J_w(\phi_n) = -\mathbb{E}_{(s,a,c,s') \sim \mathcal{D}}[\log \mathcal{M}_{\phi_n}(s',c|s,a)],
\end{equation}
\noindent where $\mathcal{D}$ denotes the real experience replay buffer. Ensemble diversity is encouraged via random initialization and stochastic mini-batch sampling.

\subsubsection{K-Level Reasoning}

To model multi-agent interactions in traffic scenarios, we incorporate K-level reasoning inspired by game theory. This hierarchical mechanism allows each agent to recursively reason about others' decisions. At reasoning level $k$, the agent predicts the behaviors of opponents acting with $(k{-}1)$-level reasoning.

At Level-0, the world model predicts the next state $\hat{s}^0$ and cost $\hat{c}^0$ based solely on the current state-action pair $(s,a)$:
\begin{equation}
\hat{s}^0, \hat{c}^0 \leftarrow \mathcal{M}^0_{\phi_n}(s,a) = \mathcal{N}(\mu^0_{\phi_n}(s,a), (\sigma^0_{\phi_n}(s,a))^2).
\end{equation}
At Level-1, the model incorporates Level-0 predictions to account for interactive effects:
\begin{equation}
\begin{aligned}
\hat{s}^1, \hat{c}^1 & \leftarrow \mathcal{M}^1_{\phi_n}(s,a,\hat{c}^0,\hat{s}^0) \\
&= \mathcal{N}(\mu^1_{\phi_n}(s,a,\hat{c}^0,\hat{s}^0),(\sigma^1_{\phi_n}(s,a,\hat{c}^0,\hat{s}^0))^2),
\end{aligned}
\end{equation}
In general, at Level-$k$, the model recursively builds on the previous level's predictions:
\begin{equation}\label{Eq_k_level}
\begin{aligned}
s', c &\leftarrow \mathcal{M}^k_{\phi_n}(s,a,\hat{c}^{k-1},\hat{s}^{k-1}) \\
&= \mathcal{N}(\mu^k_{\phi_n}(s,a,\hat{c}^{k-1},\hat{s}^{k-1}), (\sigma^k_{\phi_n}(s,a,\hat{c}^{k-1},\hat{s}^{k-1}))^2),
\end{aligned}
\end{equation}
\noindent where $(\hat{s}^{k-1}, \hat{c}^{k-1})$ are predictions from Level-$(k{-}1)$. This recursive formulation enables the model to reason about multi-agent interactions with increasing depth and fidelity.

\subsubsection{Potential Risk Modeling}

Risk is quantified by integrating predicted cost with total uncertainty. The potential risk function is defined as:
\begin{equation}\label{potential_risk}
f(s,a) = \beta \cdot \hat{c}(s,a) + (1-\beta) \cdot \hat{\sigma}(s,a),
\end{equation}
\noindent where $\beta \in [0,1]$ balances the influence of predicted cost $\hat{c}(s,a)$ and total uncertainty $\hat{\sigma}(s,a)$, both normalized to $[0,1]$.

We model total uncertainty $\hat{\sigma}(s,a)$ as the combination of two sources: 1) Aleatoric uncertainty, denoted as $\sigma^2_{\text{a}}$, accounts for the inherent randomness in the environment and is estimated directly from the ensemble model outputs, i.e., $\sigma^2_{\text{a}} = (\sigma^k_{\phi_n})^2$. And 2) epistemic uncertainty, $\sigma^2_{\text{e}}$, reflects the model's knowledge limitations and is quantified as the variance across the ensemble predictions:
\begin{equation}
\sigma^2_{\text{e}} = \mathrm{Var}(\mu^k_{\phi_n}), \quad n \in \{0,1,\ldots,N\},\; k \in \{0,1,\ldots,K\}.
\end{equation}
The total uncertainty is then computed as:
\begin{equation}\label{total_uncertainty}
\hat{\sigma}(s,a) = \kappa \cdot \sqrt{\sigma^2_{\text{a}} + \sigma^2_{\text{e}}},
\end{equation}
where $\kappa$ is a dynamic amplification coefficient modulated by interaction context:
\begin{equation}
\kappa = \min\left(u_1 \cdot \max(|v_{\text{ego}} - v_{\text{nearby}}|), u_1\right),
\end{equation}
where, $u_1$ is a manually designed maximum uncertainty amplification coefficient, and $v_{\text{ego}}$ and $v_{\text{nearby}}$ denote the velocities of the ego and nearby vehicles, respectively. This mechanism increases uncertainty in situations with high speed differences or irregular behaviors that indicate potential conflicts, enabling the model to better estimate risks under complex traffic conditions.

\subsubsection{Adaptive Prediction Horizon}

To enhance planning reliability, we introduce an adaptive prediction horizon mechanism that dynamically adjusts rollout depth $m$ based on model confidence. This approach curtails long-horizon rollouts under high uncertainty, thereby mitigating compounding errors.

Given ensemble uncertainty $\hat{\sigma}(s,a)$, the adaptive rollout length $m^*$ is computed as:
\begin{equation}\label{adaptive_rollout}
m^* = \lfloor \text{clip}(-u_2 \cdot \hat{\sigma}(s,a) + m_\text{base}, m_\text{min}, m_\text{max}) \rfloor,
\end{equation}
\noindent where $u_2$ controls sensitivity to uncertainty, $m_\text{base}$ is the nominal horizon, and $\text{clip}(\cdot)$ ensures $m^*$ remains within $[m_\text{min}, m_\text{max}]$. The floor function $\lfloor \cdot \rfloor$ ensures $m^*$ is an integer.

This mechanism shortens the prediction horizon in high-uncertainty situations--such as when the model faces unfamiliar state-action pairs--to limit error accumulation. In contrast, the horizon is extended in low-uncertainty regions to support deeper and more efficient planning. By adaptively adjusting the prediction depth, this strategy effectively balances computational efficiency and reliability in dynamic and uncertain environments.

\subsubsection{Integrated Risk Prediction}

The comprehensive potential risk model integrates predictions across all ensemble models and reasoning levels. Following $m$ steps of lookahead planning, the potential risk is computed as:
\begin{equation}\label{Eq_risk_model}
f(s^m,a^m) = \beta \cdot \hat{c}(s^m,a^m) + (1-\beta) \cdot \hat{\sigma}(s^m,a^m),
\end{equation}
\noindent where $s^m$ and $a^m$ are the state and action after $m$ planning steps. The ensemble predictions are aggregated as:
\begin{equation}
(s',\hat{c}) \sim \frac{1}{N}\sum_{k=1}^N \mathcal{M}_{\phi_n}(s,a), \quad \hat{\sigma} = \gamma \cdot \frac{1}{N}\sum_{k=1}^N \sigma_{\phi_n}(s,a).
\end{equation}
This framework combines K-level reasoning, dual uncertainty quantification, and context-aware amplification, enabling robust, risk-sensitive decision-making for AD in dynamic environments.

\subsection{Reachability Modeling}

\subsubsection{Barrier Scaling Factor}

To enhance reachability modeling under dynamic and uncertain conditions, we introduce an auxiliary neural network termed the barrier policy $\mathcal{B}$, parameterized by $\psi$. Given the current state $s$, this network outputs a scalar barrier factor $b \in [0, 1]$, which modulates the strength of the reachability constraints in a dynamic and adaptive manner. A higher value of $b$ imposes more conservative (i.e., tighter) constraints, allowing the model to adaptively adjust the effective safety boundaries around the ego vehicle in response to varying risk levels. This adaptive mechanism enhances both safety and reachability, allowing the system to balance conservatism and maneuverability in uncertain environments.

\subsubsection{Barrier-Guided Reachability}
The barrier policy $\mathcal{B}$ is integrated into the world model to adaptively adjust reachability constraints based on predicted uncertainty $\hat{\sigma}(s, a)$. For each surrounding vehicle $i \in \mathcal{I}_\text{obstacle}$, the adjusted relative distance in the next state, denoted as $\tilde{d}'_i$, is computed as:
\begin{equation}\label{reachability_constrain_1}
\tilde{d}'_i =
\begin{cases} 
\max(d'_i - \bar{b}, 0), & \text{if } i \in \mathcal{I}_\text{obstacle}, \\
d'_i, & \text{otherwise,}
\end{cases}
\end{equation}
\noindent where $\bar{b}$ is an uncertainty-based adjustment term:
\begin{equation}\label{reachability_constrain_2}
\bar{b} = b \cdot \hat{\sigma}(s, a) \cdot \text{abs}(\mathcal{N}(0, 1)) \cdot u_3,
\end{equation}
\noindent where $\mathcal{N}(0, 1)$ is a standard Gaussian random variable, and $u_3$ is a tunable scaling factor.

\subsubsection{Learning objective} 
The barrier policy $\mathcal{B}$ is trained to minimize the potential risk function $f(s, a)$, as defined in Eq.(\ref{potential_risk}). The corresponding training loss is formulated as:
\begin{equation}\label{update_bar}
\mathcal{L}_\text{bar}(\psi) = \mathbb{E}_{s \sim \mathcal{D} \cup \mathcal{D}'} \big[f(s, a)\big],
\end{equation}
\noindent where $\mathcal{D}'$ is the virtual experience buffer generated by the world model. 

During both training and trajectory rollout, the learned barrier policy dynamically adjusts reachability constraints in real time. By adapting the effective safety margin to contextual risk and uncertainty at each step, the ego vehicle is guided to generate feasible and safe trajectories, significantly enhancing adaptability and robustness in complex, dynamic environments.

\subsection{Potential Risk-Constrained RL}

This section introduces a constrained RL framework aimed at optimizing driving strategies while maintaining bounded risk levels.

Under the CMDP framework, GTR2L seeks to solve the following constrained maximization problem:
\begin{equation}\label{Eq_optimization_problem}
\begin{aligned}
\max_{\pi} \mathbb{E} \left[ \sum_{t=0}^{\infty} \gamma^t r(s_t, a_t) \right], \quad
\text{s.t.} \quad \mathbb{E} [f(s^m, a^m)] \leq f_0,
\end{aligned}
\end{equation}
with temporal index $t$ and predefined boundary $f_0$.

To solve this, we adopt a risk-constrained policy iteration method, which alternates between policy evaluation and improvement phases. The constrained problem is reformulated in Lagrangian form:
\begin{equation}\label{Eq_Lagrange_duality_theory}
L(\pi, \lambda) = \mathbb{E}\left[ \sum_{t=0}^{\infty} \gamma^t r(s_t, a_t) + \lambda (f_0 - f(s^m, a^m)) \right],
\end{equation}
with non-negative dual variable $\lambda$.

\begin{figure*}
    \begin{center}
    \includegraphics[width=0.8\linewidth]{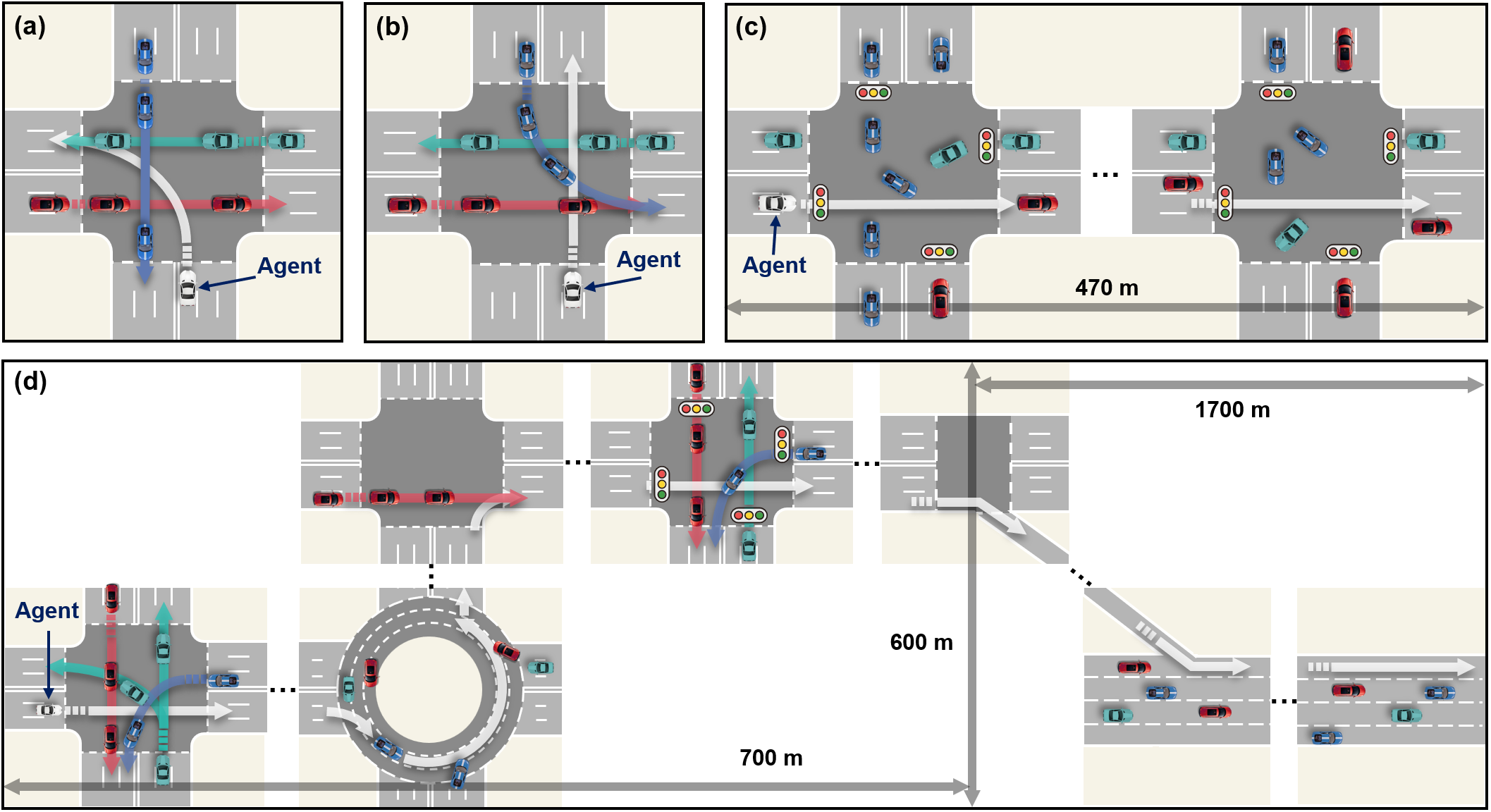}
    \caption{Regular traffic scenarios: (a) unprotected left-turn at an unsignalized intersection (Scenario~1), (b) unprotected straight-crossing at an unsignalized intersection (Scenario~2), (c) consecutive signalized intersections with mixed traffic flows (Scenario~3), and (d) long-distance composite scenario with mixed traffic flows (Scenario~4).} 
    \label{sumo_scenarios}
 \end{center}                                       
\end{figure*}

\subsubsection{Risk-Constrained Policy Evaluation}

The action-value function $Q^{\pi}(s, a)$ is refined using the Bellman operator $\mathcal{T}$:
\begin{equation}
\mathcal{T}Q^{\pi}(s, a) \equiv r(s, a) + \gamma \; \mathbb{E}_{s' \sim p} [V^{\pi}(s')],
\end{equation}
where the value function $V^{\pi}(\cdot)$ is defined as:
\begin{equation}
V^{\pi}(s) = \mathbb{E}_{a \sim \pi(\cdot|s)} [Q^{\pi}(s, a) - \lambda f(s^m, a^m)].
\end{equation}

To stabilize learning, we employ a pair of utility networks parameterized by $\phi^z$, where $z \in \{1, 2\}$. These are trained by minimizing the critic loss:
\begin{equation}\label{loss_critic}
J_c(\phi^z) = \mathbb{E}_{(s,a,r,s') \sim \mathcal{M}} \left[ \|y - Q^{\pi}(s, a; \phi^z)\|^2_2 \right],
\end{equation}
where $y$ denotes the target Q-value. Following empirical findings~\cite{janner2019trust}, we exclusively train utility networks on real interaction data, reducing reliance on model-generated samples and improving training quality.

To ensure safety, the Q-values for unsafe actions must be strictly lower than those of safe ones. Following the horizon-limited safety condition~\cite{thomas2021safe, he2023fear}, a penalty coefficient $c^*$ is defined to ensure that unsafe policies receive reduced returns beyond a certain planning horizon $H$:
\begin{equation}
c^* = \frac{\bar{r} - \underline{r}}{\gamma^H(1 - \gamma)} + \frac{\lambda}{\gamma^H} - \frac{\bar{r}}{1 - \gamma},
\end{equation}
where $\bar{r}$ and $\underline{r}$ are the upper and lower bounds of the reward. This penalty guides the agent to avoid unsafe actions and prioritize safety during policy optimization.

To mitigate value overestimation, we use the minimum of the dual target utility networks for critic updates. The target value $y$ is computed as:
\begin{equation}
y = 
\begin{cases}
r + \gamma \tilde{Q}^{\pi}(s', a'), & \text{if safety}, \\
-c^*, & \text{else},
\end{cases}
\end{equation}
where $a'$ sampled from the policy $\pi(\cdot|s')$. And the target value function is defined as:
\begin{equation}
\tilde{Q}^{\pi}(s, a) = \min_{z \in \{1,2\}} \hat{Q}^{\pi}(s, a; \bar{\phi}^z) - \lambda f(s^m, a^m),
\end{equation}
where $\bar{\phi}^z$ denotes the target network parameters for $z \in \{1, 2\}$, and $f(s^m, a^m)$ is the potential risk after $m$ planning steps. Target value function parameters $\phi^z$ update through exponential averaging: $\bar{\phi}^z \leftarrow \tau \bar{\phi}^z + (1 - \tau) \phi^z$, Polyak averaging coefficient $\tau \in (0,1)$.

\subsubsection{Risk-Constrained Policy Improvement}

Policy improvement aims to maximize expected return while satisfying the safety constraint. Applying Lagrangian duality theory with Eq.(\ref{Eq_Lagrange_duality_theory}), constrained optimization from Eq.(\ref{Eq_optimization_problem}) transforms into:
\begin{equation}
\begin{aligned}
\min_{\lambda} \max_{\pi} L(\pi, \lambda) = \min_{\lambda} \max_{\pi} \mathbb{E} \left[ \sum_{t=0}^{\infty} \gamma^t r(s_t, a_t) \right. \\
\left. + \lambda (f_0 - f(s^m, a^m)) \right].
\end{aligned}
\end{equation}

To ensure robustness and behavioral diversity, we optimize the actor using both real and model-generated data. The actor loss is defined as:
\begin{equation}\label{loss_actor}
J_a(\theta) = \mathbb{E}_{s \sim \mathcal{D} \cup \mathcal{D}', a \sim \pi(\cdot|s;\theta)} [Q^{\pi}(s, a) - \lambda f(s^m, a^m)].
\end{equation}

Simultaneously, dual variable $\lambda$ is updated via:
\begin{equation}\label{update_dual}
J_d(\lambda) = \mathbb{E}_{s \sim \mathcal{D} \cup \mathcal{D}'} [\lambda (f_0 - f(s^m, a^m))].
\end{equation}

Our implementation interprets expense $\hat{c}$ from world models as safety violation probability. During evaluation phases, risk mitigation involves safety assessment using trained world models. When actions receive high collision risk ratings from world models, Gaussian noise $\epsilon$ modifies those actions appropriately (illustrated in Fig.~\ref{framework}). The detailed procedure of the proposed GTR2L algorithm is summarized in Algorithm~\ref{alg:GTR2L}.

\begin{algorithm}[t]
\caption{Game-Theoretic Risk-Shaped Reinforcement Learning (GTR2L)}
\label{alg:GTR2L}
\begin{algorithmic}[1]
\State \textbf{Initialize:}
\State \quad Ensemble world model parameters $\{\phi_n^k\}$, Barrier parameters $\psi$, Actor parameters $\theta$, Dual variable $\lambda$
\State \quad Critic parameters $\phi^1,\phi^2$ and targets $\bar{\phi}^1, \bar{\phi}^2$
\State \quad Real and virtual experience buffer $\mathcal{D}$, $\mathcal{D}'$
\For{episode $n_e = 1$ to $N_e$}
    \State Reset initial state $s_0$
    \For{timestep $t=1$ to $T$}
        \State Execute $a_t \sim \pi(\cdot|s_t;\theta)$; observe $s_{t+1}, r_t, c_t$
        \State Store $(s_t, a_t, r_t, c_t, s_{t+1})$ in $\mathcal{D}$
    \EndFor
    \For{world model update step}
        \State Sample mini-batch from $\mathcal{D}$
        \For{each model $n$ and level $k$}
            \State Update world model $\phi_n^k$ using Eq.~(\ref{update_WM})
        \EndFor
    \EndFor
    \For{agent update step}
        \State Sample states $s$ from $\mathcal{D}$
        \State Estimate uncertainty $\hat{\sigma}(s,a)$ via Eq.~(\ref{total_uncertainty})
        \State Adapt horizon $m^*$ via Eq.~(\ref{adaptive_rollout})
        \For{$m=1$ to $m^*$}
            \State Sample $\hat{a}^m \sim \pi(\cdot|\hat{s}^{m};\theta)$
            \State Use K-level model to rollout $(\hat{s}^m, \hat{c}^m)$ (Eq.~(\ref{Eq_k_level}))
            \State Compute potential risk $f(\hat{s}^m, \hat{a}^m)$ (Eq.~(\ref{Eq_risk_model}))
            \State Update reachability boundary (Eq.~(\ref{reachability_constrain_1},~\ref{reachability_constrain_2}))
            \State Store $(\hat{s}^{m}, \hat{a}^m, \hat{c}^m, \hat{s}^{m+1})$ in $\mathcal{D}'$
        \EndFor
        \State Update critic $\phi^1,\phi^2$ using samples from $\mathcal{D}$ (Eq.~(\ref{loss_critic}))
        \State Update actor $\theta$ using $\mathcal{D} \cup \mathcal{D}'$ (Eq.~(\ref{loss_actor}))
        \State Update barrier $\psi$ (Eq.~(\ref{update_bar})), dual variable $\lambda$ (Eq.~(\ref{update_dual})) using samples from $\mathcal{D} \cup \mathcal{D}'$
        \State Update targets via Polyak averaging
    \EndFor
\EndFor
\end{algorithmic}
\end{algorithm}

\section{Experimental Validation}\label{Sec_Experimental}

\subsection{Driving Scenarios}

To facilitate clear reference and comparison, we assign indices to each scenario. The four simulation of urban mobility (SUMO)-based urban scenarios are denoted as Scenario~1-4, while the two Car learning to act (CARLA)-based safety-critical scenarios are referred to as Scenario~5-6, respectively. This unified naming convention simplifies cross-scenario analysis throughout the experimental validation.

\subsubsection{Regular Traffic Scenarios}

The following four representative scenarios, denoted as Scenario~1-4, are constructed within the SUMO platform to evaluate agent decision-making and behavioral performance in complex urban settings. Each scenario targets a specific aspect of urban traffic complexity:

- Scenario 1: Unprotected left-turn: In the scenario illustrated in Fig.~\ref{sumo_scenarios}(a), the agent is required to execute an unprotected left turn at a four-way intersection, where it must interact with both oncoming traffic and lateral vehicle flows. This scenario tests the agent’s ability to handle complex right-of-way negotiations without traffic signals.

- Scenario 2: Unprotected straight-crossing: As shown in Fig.~\ref{sumo_scenarios}(b), the agent drives straight through an unsignalized intersection, dynamically negotiating with oncoming left-turn vehicles and lateral traffic. This emphasizes real-time interaction and collision avoidance in ambiguous priority situations.

- Scenario 3: Consecutive signalized intersections: Fig.~\ref{sumo_scenarios}(c) presents a scenario with two consecutive signalized intersections spaced 470 meters apart. The agent is required to follow a predetermined path (white arrow), maintaining stable speed and avoiding interference from vehicles approaching from all directions. This tests the agent’s ability to obey traffic signals and manage speed control across multiple intersections.

- Scenario 4: Long-distance composite scenario: In the comprehensive scenario depicted in Fig.~\ref{sumo_scenarios}(d), various elements are integrated, including unsignalized intersections, roundabouts, signalized intersections, and highway driving. The agent proceeds straight through an unsignalized intersection, passes through a roundabout and another unsignalized intersection, then turns right at a signalized intersection. Subsequently, it enters the highway via an on-ramp, and completes a cruising task. This comprehensively evaluates the global planning and adaptability of agents under different multi-stage transportation conditions.

\begin{figure}
    \begin{center}
    \includegraphics[width=0.85\linewidth]{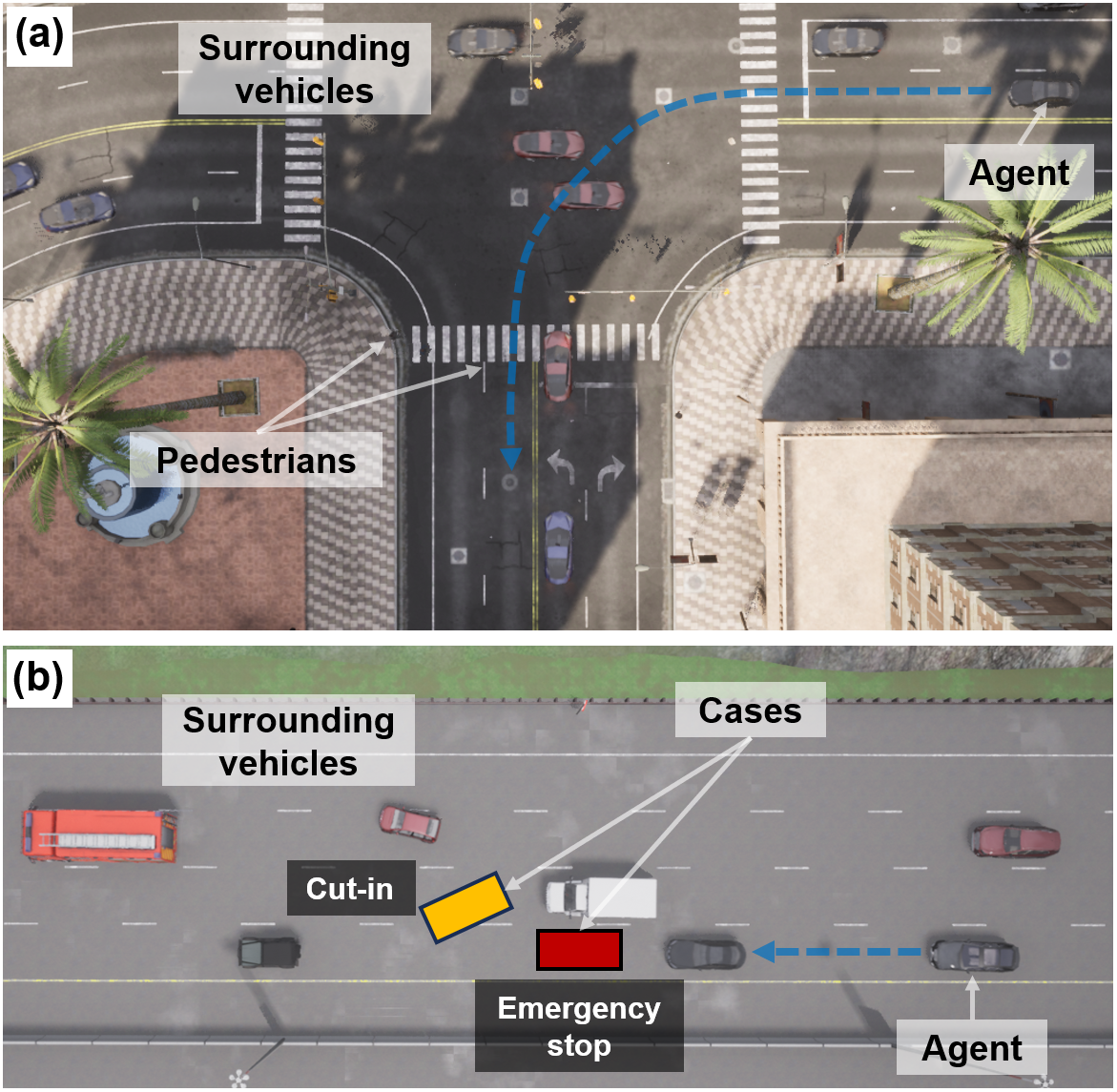}
    \caption{Safety-critical scenarios: (a) unprotected left-turn with pedestrian crossing at an urban intersection (Scenario~5), and (b) highway scenario with random emergency events (Scenario~6), where either a sudden emergency stop or a sudden cut-in occurs randomly in each episode.} \label{carla_scenarios}
 \end{center}                                       
\end{figure}

\subsubsection{Safety-Critical Scenarios}

Two core safety-critical scenarios, denoted as Scenario~5 and Scenario~6, are designed in the high-fidelity CARLA simulator to assess agent decision-making and emergency response under extreme conditions (as shown in Fig.~\ref{carla_scenarios}). These scenarios are constructed in CARLA, as conducting such experiments in the real world is prohibitively dangerous. Using CARLA ensures both safety and realism when evaluating agent behavior in rare or hazardous traffic situations.

- Scenario 5: Unprotected left-turn with pedestrian crossing: In this scenario, the agent must perform an unprotected left turn at a busy urban intersection, surrounded by 10 to 15 vehicles, each initialized with a random speed between 20 and 50~km/h. Pedestrians, each walking at a random speed between 1 and 1.5~m/s, actively cross the street along the agent’s intended path. The agent is required to monitor both oncoming vehicles and crossing pedestrians, yielding appropriately to ensure the safety of all road users.

- Scenario 6: Highway scenario with emergency cases: This scenario takes place on a multi-lane highway, where the agent cruises among 15 to 25 surrounding vehicles. Each vehicle is initialized with a random target speed between 40 and 70 km/h. During each episode, the environment randomly triggers rare but critical emergencies, including: a) Sudden Emergency Stop--where the lead vehicle may abruptly decelerate to a stop without warning, requiring the agent to promptly detect the hazard and take evasive action; and b) Sudden Cut-in--where an adjacent vehicle may quickly change lanes and cut in front of the agent, reducing the safety margin and requiring rapid adjustment of trajectory and speed to maintain safety.

\begin{figure}
    \begin{center}
    \includegraphics[width=0.85\linewidth]{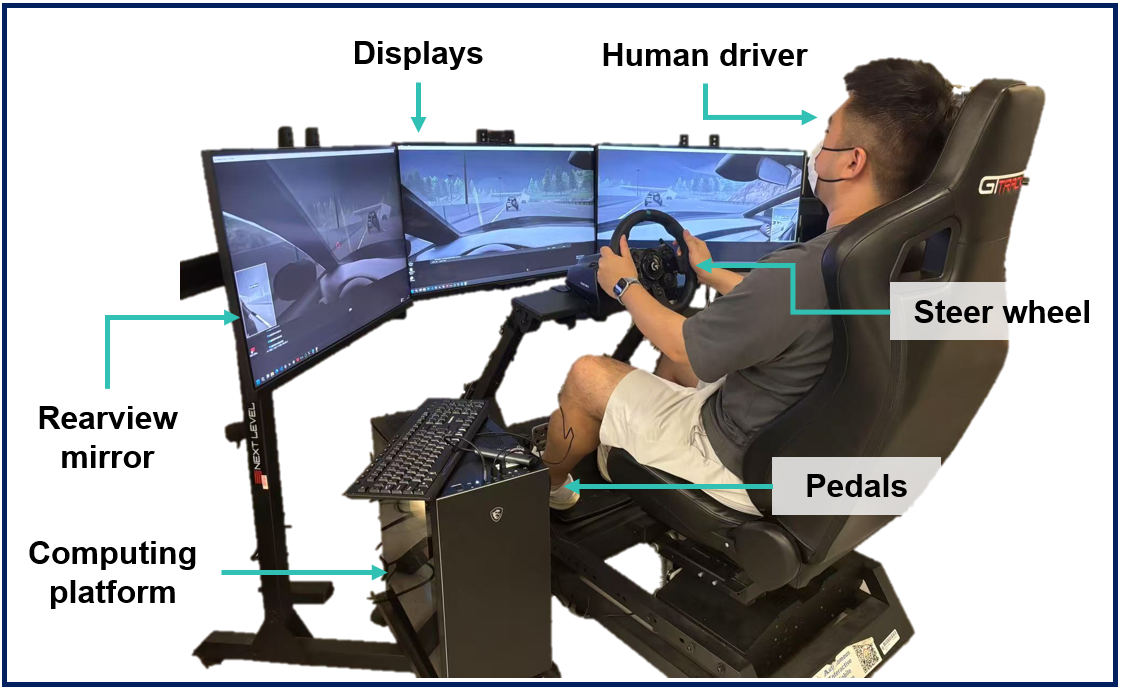}
    \caption{The experimental driving simulator platform is specifically designed for the evaluation of safety-critical scenarios. Within this environment, a human participant operates the ego vehicle through the use of a steering wheel and pedal set. The system comprises a dedicated computing unit and three high-resolution monitors, offering high-fidelity visual feedback to ensure an immersive and realistic driving experience during testing.} \label{hil_platform}
 \end{center}                                       
\end{figure}

\subsection{RL Problem Definition}

\subsubsection{State Space}

In all scenarios (Scenarios~1-6), the agent’s state representation includes information about the ego vehicle and its six nearest neighbors within a 200-meter radius, categorized as front, rear, front-left, rear-left, front-right, and rear-right. For each neighboring vehicle $i$, the state captures relative distance $d_i$, relative direction $\phi_i$, speed $v_i$, and heading angle $\theta_i$. The ego vehicle’s speed $v_\text{ego}$ and heading direction $\theta_\text{ego}$ are also included, resulting in a vehicle state representation:
\begin{equation}
\mathcal{S}_{\text{vehicle}} = \bigcup_{i=1}^6 (d_i, \phi_i, v_i, \theta_i) \cup (v_{\text{ego}}, \theta_{\text{ego}}),
\end{equation}
with a total vehicle-related dimension of 26.

For Scenario~3-4, and Scenario~5-6, the state further incorporates goal-related information:
\begin{equation}
\mathcal{S}_{\text{goal}} = ( d_\text{goal} ),
\end{equation}
where $d_\text{goal}$ is the distance from the ego vehicle to the navigation goal.

In scenarios with traffic lights (Scenario~3-4 in SUMO), the state is extended to include traffic light features:
\begin{equation}
\mathcal{S}_{\text{road}} = ( d_\text{tl}, z_\text{tl}),
\end{equation}
where $d_\text{tl}$ is the distance to the nearest traffic light and $z_\text{tl}$ is its current phase.

\subsubsection{Action Space}

In Scenario~1-2, the agent's action is a continuous scalar representing longitudinal acceleration or deceleration, $acc \in \mathbb{R}$, where $acc > 0$ indicates acceleration and $acc < 0$ indicates deceleration.

In Scenario~3-4, and Scenario~5-6, the action space is expanded to include both acceleration and lane-change maneuvers:
\begin{equation}
\mathcal{A} = [acc, L],
\end{equation}
where $acc$ denotes the longitudinal acceleration and $L$ represents the lane-change decision (left, none, or right).

\subsubsection{Reward and Cost Function}
The reward and cost functions are designed to balance efficiency and safety across all scenarios. The basic reward encourages efficient driving through the ego vehicle's speed:
\begin{equation}
r_{\text {v}}(s,a) = \frac{v_\text{ego}}{\omega_1},
\end{equation}
\noindent where $\omega_1 = 10$ for Scenario~1-2, and $\omega_1 = 5$ for all other scenarios. 

For Scenario~3-4, and Scenario~5-6, an additional navigation reward is provided for reaching the goal:
\begin{equation}
r(s,a) = r_{\text {v}} + 
\begin{cases} 
100.0, & \text{if reaching the goal,} \\
-\log\left(1.0 + \frac{d_\text{goal}}{d_\text{max}}\right), & \text{otherwise.}
\end{cases}
\end{equation}
\noindent Here, $d_\text{max}$ is the maximum possible distance.

The cost function penalizes unsafe behaviors, such as collisions and traffic violations:
\begin{equation}
c(s,a) = c_\text{collision} + c_\text{violation},
\end{equation}
\noindent where $c_\text{collision} = 1.0$ if a collision occurs (otherwise 0), and $c_\text{violation} = 1.0$ if the agent runs a red light (in Scenario~3-4), otherwise 0. The final reward function is calculated as: $\tilde{r}(s,a) = r(s,a) - c(s,a)$.

\subsection{Baseline Methods}

To evaluate the effectiveness of the proposed method, several representative baselines are selected for comparison:

\subsubsection{Rule-based Method}
The intelligent driver model (IDM) is a classic car-following rule-based method that calculates acceleration based on speed, distance, and relative velocity to the lead vehicle. In our experiments, IDM is combined with a fixed lane-following rule to control the agent in all scenarios.

\subsubsection{SAC}
SAC~\cite{haarnoja2018soft} is an advanced off-policy RL algorithm that optimizes both the expected return and entropy, enabling efficient exploration in continuous action spaces.

\subsubsection{SAC-Lag}
SAC-Lag~\cite{ha2020learning} is an extension of SAC that incorporates Lagrangian relaxation to handle safety constraints during training. This ensures that the learned policy not only pursues high rewards but also adheres to predefined safety requirements.

\subsubsection{SMBPO}
Safe model-based policy optimization (SMBPO)~\cite{thomas2021safe} uses a world model to predict short-term states and penalize unsafe trajectories, providing theoretical guarantees to avoid dangerous states under certain assumptions.

\subsubsection{FNI-RL}
Fear-neuro-inspired RL (FNI-RL)~\cite{he2023fear} is a state-of-the-art model-based safe RL algorithm specifically designed to enhance the performance of AD systems. Inspired by the amygdala's defensive mechanism in the human brain, FNI-RL enables agents to proactively learn defensive driving behaviors and effectively reduce safety violations.

\subsubsection{Human Driver}
To provide a practical baseline, 10 licensed human drivers operate the ego vehicle on the simulator platform shown in Fig.~\ref{hil_platform}. All drivers underwent a standardized training session to ensure consistent familiarity with the simulator and task requirements before testing. Providing a practical baseline for safety and driving competence comparison.

All baselines were trained and evaluated under identical settings as the proposed method, with carefully tuned hyperparameters to ensure fair comparison.

\begin{figure*} [t!]
\centering
    \subfloat[]{
        \includegraphics[width=0.28\linewidth]{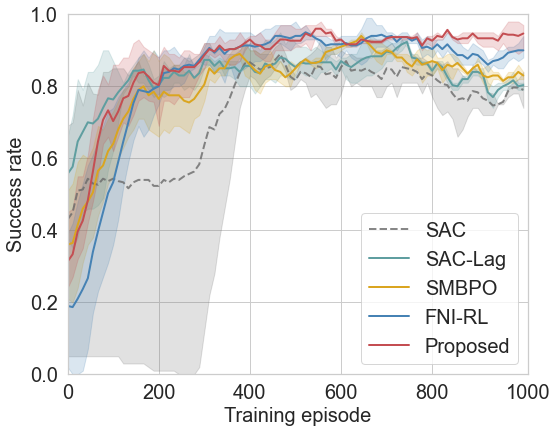}}
    \subfloat[]{
        \includegraphics[width=0.28\linewidth]{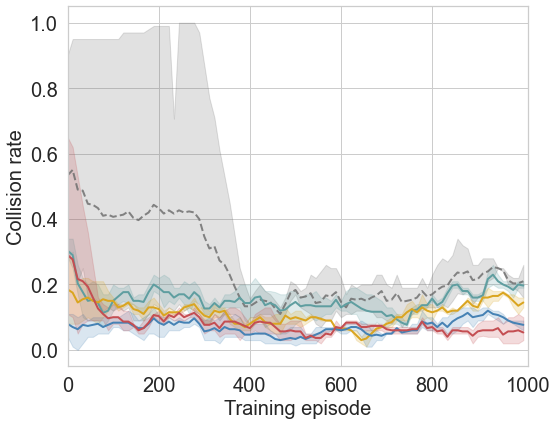}}
    \subfloat[]{
        \includegraphics[width=0.285\linewidth]{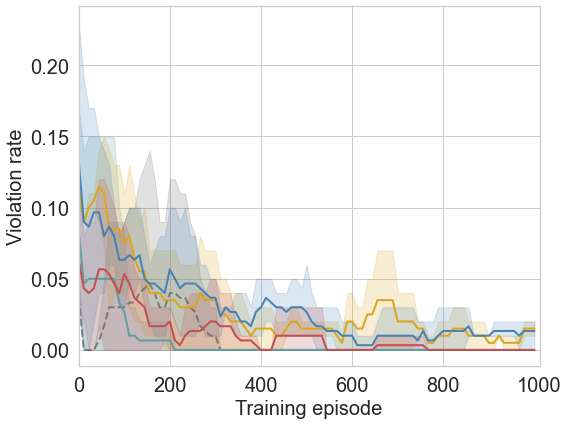}} \\
    \subfloat[]{
        \includegraphics[width=0.28\linewidth]{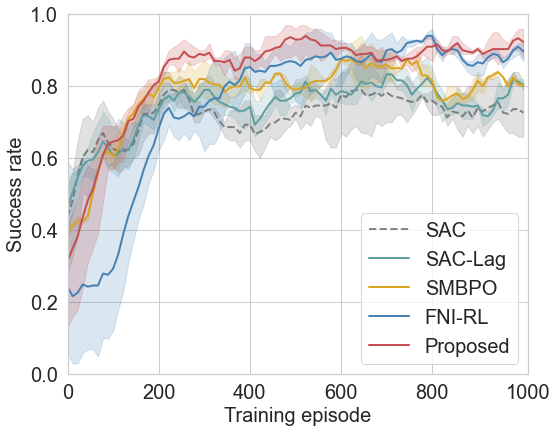}}
    \subfloat[]{
        \includegraphics[width=0.28\linewidth]{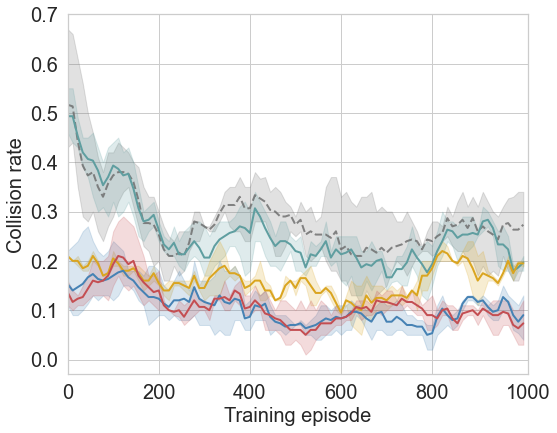}}
    \subfloat[]{
        \includegraphics[width=0.285\linewidth]{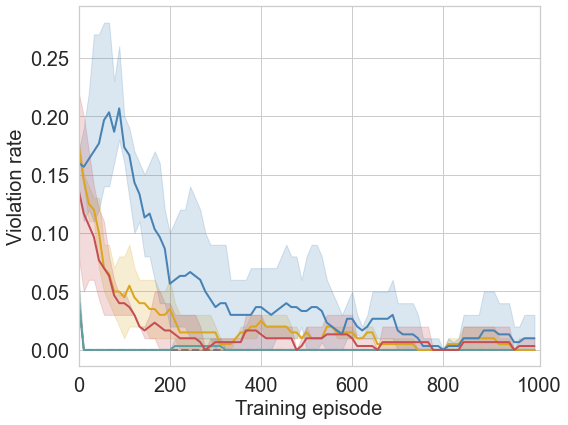}} 
    \caption{Training performance of different autonomous driving agents on the Scenario 3 with traffic signals under varying traffic flow densities. Flow-1: (a) Success rate, (b) Collision rate, (c) Traffic-light violation rate; Flow-2: (e) Success rate, (f) Collision rate, (g) Traffic-light violation rate.}
    \label{training_env8} 
\end{figure*}

\begin{figure*} [t!]
\centering
    \subfloat[]{
        \includegraphics[width=0.28\linewidth]{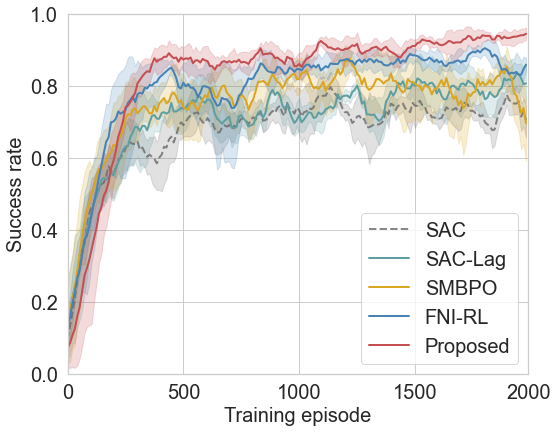}}
    \subfloat[]{
        \includegraphics[width=0.28\linewidth]{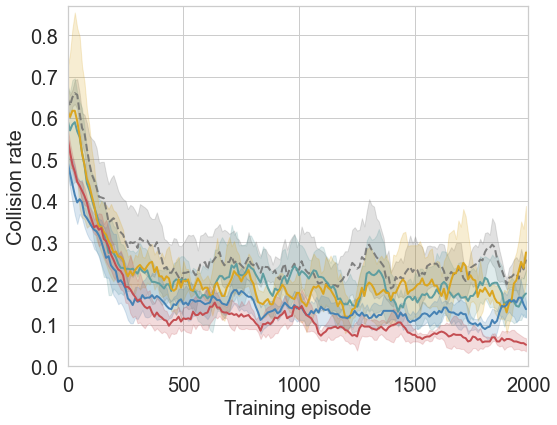}}
    \subfloat[]{
        \includegraphics[width=0.285\linewidth]{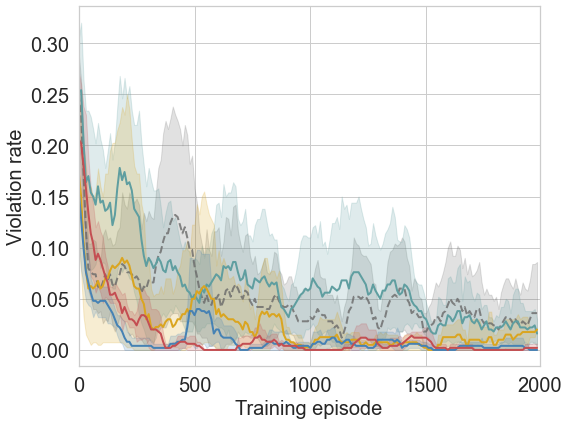}} 
    \caption{Training performance of different agents on the Scenario 4 under stochastic and dynamic traffic conditions. (a) Success rate, (b) Collision rate, (c) Traffic-light violation rate.}
    \label{training_env6} 
\end{figure*}

\subsection{Implementation Details}

All agents were trained using five different random seeds. For Scenario~4 (long-distance composite scenario), each agent was trained for 2000 episodes, while 1000 episodes of training were performed for all other scenarios (Scenario~1-3 and Scenario~5-6). 

To comprehensively evaluate performance, two traffic flow densities, flow-1 and flow-2, were configured in the SUMO environment. For Scenario~1-3, flow-1 and flow-2 were used separately for training and evaluation, where vehicles were generated with probabilities of 0.5 and 0.8 per second, respectively. For Scenario~4, flow-1 and flow-2 were mixed during both training and evaluation to further increase traffic diversity. All vehicles were equipped with built-in AI agents exhibiting stochastic behaviors.

For the CARLA-based scenarios (Scenario~5 and Scenario~6), the environment was similarly adapted for diversity and realism. The agent's start position was randomly assigned along the initial route, and the goal was set as the end of a selected route. To ensure realistic simulations, pedestrian participants were generated at random times, and all traffic vehicles were managed by the built-in traffic manager, leading to a wide variety of randomly chosen driving behaviors.

For model evaluation, we systematically tested the final policies obtained from each algorithm across all random seeds. To ensure a fair comparison, every method employed an identical policy network architecture throughout the experiments. Relevant hyperparameters of proposed method are summarized in Table~\ref{table:hyperparameters}.

\begin{table}[htbp]
\renewcommand{\arraystretch}{1.3}
\centering
\caption{Hyperparameter Setting of Proposed Method.}
\label{table:hyperparameters}
\begin{tabular}{ccc}
\hline
Symbol & Definition & Value \\
\hline
$l_a$ & Actor learning rate & 3e$^{-4}$ \\
$l_c$ & Critic learning rate & 3e$^{-4}$ \\
$l_d$ & Dual learning rate & 1e$^{-3}$ \\
$l_w$ & World model learning rate & 1e$^{-3}$ \\
$l_b$ & Barrier learning rate & 1e$^{-3}$ \\
$\tau$ & Polyak averaging coefficient & 0.995 \\
$\gamma$ & Discount factor & 0.99 \\
$f_0$ & Constraint threshold & 0.5 \\
$\beta$ & Risk model's weight & 0.90 \\
$-$ & RL batch size & 64 \\
$-$ & World model batch size & 128 \\
\hline
\end{tabular}
\end{table}

\begin{table*}[t]
\centering
\renewcommand{\arraystretch}{1.3}
\caption{Statistical Results of Different Autonomous Driving Agents in the SUMO Scenarios (a)-(d), Including the Mean and Standard Deviation (In Brackets)}
\label{tab:test_somo}
\begin{tabular}{cccccccccc}
\hline
\multirow{2}{*}{Method} & \multirow{2}{*}{Metric} & \multicolumn{2}{c}{Scenario 1} & \multicolumn{2}{c}{Scenario 2} & \multicolumn{2}{c}{Scenario 3} & \multicolumn{1}{c}{Scenario 4} & \multirow{2}{*}{Summary} \\
\cline{3-9}
& & Flow-1 & Flow-2 & Flow-1 & Flow-2 & Flow-1 & Flow-2 & Dynamic & \\
\hline
\multirow{5}{*}{SAC} & DS & 0.96 (0.01) & 0.91 (0.05) & 0.99 (0.01) & 0.98 (0.01) & 0.96 (0.07) & 0.87 (0.09) & 0.72 (0.12) & 0.91 (0.05) \\
                     & SR & 0.99 (0.01) & 0.92 (0.07) & 0.99 (0.01) & 0.98 (0.01) & 0.79 (0.11) & 0.72 (0.12) & 0.72 (0.14) & 0.87 (0.07) \\
                     & CR & 0.01 (0.01) & 0.08 (0.07) & 0.01 (0.01) & 0.02 (0.01) & 0.21 (0.11) & 0.28 (0.12) & 0.25 (0.13) & 0.12 (0.07)  \\
                     & VR & - -&- -&- -&- -                                       & 0.00 (0.00) & 0.00 (0.00) & 0.03 (0.04) & 0.01 (0.01) \\
& TNC ($\times 10^2$)     & 1.67 (0.84) & 1.10 (0.59) & 0.22 (0.11) & 0.26 (0.12) & 2.70 (1.40) & 2.97 (0.20) & 5.56 (1.06) & 2.21 (0.62) \\
\hline
\multirow{5}{*}{SAC-Lag} & DS & 0.96 (0.02) & 0.93 (0.05) & 0.99 (0.01) & 0.98 (0.00) & 0.96 (0.05) & 0.92 (0.08) & 0.78 (0.10) & 0.93 (0.04) \\
                         & SR & 0.99 (0.02) & 0.95 (0.06) & 0.99 (0.01) & 0.98 (0.00) & 0.82 (0.05) & 0.77 (0.09) & 0.80 (0.13) & 0.90 (0.05) \\
                         & CR & 0.01 (0.01) & 0.05 (0.06) & 0.01 (0.01) & 0.02 (0.00) & 0.18 (0.05) & 0.23 (0.09) & 0.18 (0.13) & 0.10 (0.05) \\
                         & VR & - -&- -&- -&- -                                       & 0.00 (0.00) & 0.00 (0.00) & 0.02 (0.03) & 0.01 (0.01) \\
& TNC ($\times 10^2$)         & 0.83 (0.16) & 1.32 (0.24) & 0.22 (0.12) & 0.33 (0.01) & 1.66 (0.16) & 2.65 (0.11) & 4.54 (1.25) & 1.65 (0.29) \\
\hline
\multirow{5}{*}{SMBPO} & DS & 0.95 (0.04) & 0.92 (0.05) & 0.96 (0.03) & 0.94 (0.03) & 0.93 (0.07) & 0.87 (0.06) & 0.73 (0.12) & 0.90 (0.06) \\
                       & SR & 0.96 (0.03) & 0.95 (0.04) & 0.98 (0.02) & 0.96 (0.03) & 0.84 (0.07) & 0.80 (0.08) & 0.77 (0.15) & 0.89 (0.06) \\
                       & CR & 0.04 (0.03) & 0.05 (0.04) & 0.02 (0.02) & 0.04 (0.03) & 0.14 (0.07) & 0.19 (0.09) & 0.21 (0.14) & 0.10 (0.06) \\
                       & VR & - -&- -&- -&- -                                       & 0.01 (0.01) & 0.01 (0.01) & 0.02 (0.02) & 0.01 (0.01) \\
& TNC ($\times 10^2$)       & 0.51 (0.17) & 0.48 (0.14) & 0.25 (0.13) & 0.27 (0.16) & 1.18 (0.19) & 1.67 (0.01) & 4.59 (0.91) & 1.28 (0.24) \\
\hline
\multirow{5}{*}{FNI-RL} & DS & 0.94 (0.03) & 0.94 (0.03) & 0.96 (0.04) & 0.97 (0.03) & 0.97 (0.06) & 0.94 (0.05) & 0.79 (0.08) & 0.93 (0.15) \\
                        & SR & 0.96 (0.04) & 0.96 (0.04) & 0.96 (0.05) & 0.97 (0.03) & 0.90 (0.07) & 0.88 (0.05) & 0.85 (0.10) & 0.92 (0.05) \\
                        & CR & 0.04 (0.03) & 0.03 (0.04) & 0.04 (0.04) & 0.03 (0.02) & 0.09 (0.07) & 0.11 (0.04) & 0.14 (0.09) & 0.07 (0.05) \\
                        & VR & - -&- -&- -&- -                                       & 0.01 (0.06) & 0.01 (0.02) & 0.00 (0.00) & \textbf{0.00 (0.03)} \\
& TNC ($\times 10^2$)        & 0.66 (0.18) & 0.50 (0.14) & 0.38 (0.19) & 0.31 (0.06) & 0.71 (0.06) & 1.09 (0.19) & 3.40 (0.29) & 1.01 (0.05) \\
\hline

\multirow{5}{*}{Proposed} & DS & 0.96 (0.03) & 0.96 (0.02) & 1.00 (0.00) & 0.99 (0.01) & 0.97 (0.03) & 0.97 (0.06) & 0.87 (0.05) & \textbf{0.96 (0.03)} \\
                          & SR & 0.98 (0.03) & 0.98 (0.03) & 1.00 (0.00) & 0.99 (0.01) & 0.95 (0.04) & 0.91 (0.07) & 0.94 (0.06) & \textbf{0.96 (0.03)} \\
                          & CR & 0.02 (0.03) & 0.02 (0.03) & 0.00 (0.00) & 0.01 (0.01) & 0.05 (0.04) & 0.09 (0.07) & 0.06 (0.06) & \textbf{0.04 (0.03)} \\
                          & VR & - -&- -&- -&- -                                       & 0.00 (0.00) & 0.01 (0.01) & 0.00 (0.00) & \textbf{0.00 (0.00)} \\
& TNC ($\times 10^2$)          & 0.73 (0.20) & 0.53 (0.11) & 0.19 (0.02) & 0.23 (0.11) & 0.94 (0.33) & 1.11 (0.54) & 2.84 (0.24) & \textbf{0.94 (0.22)} \\
\hline
\end{tabular}
\end{table*}

\subsection{Evaluation Metrics}

To quantitatively evaluate driving performance and safety, we adopt several comprehensive metrics:

First, we define a driving score (DS) to reflect overall driving quality:
\begin{equation}
    \mathrm{DS} = \eta\, \mathrm{SR} + (1 - \eta) \frac{v}{v_{\max}},
\end{equation}
where $\mathrm{SR}$ represents the success rate, $v$ is the agent's average speed, $v_{\max}$ is the maximum allowed speed, and $\eta$ is a weighting coefficient (set to 0.8 in our study). A successful episode is one in which the vehicle reaches the predefined target point without any safety violations, such as collisions or traffic signal violations. For safety evaluation, we report the collision rate (CR), which measures the probability of collisions during driving. In scenarios involving traffic lights, we also compute the violation rate (VR), representing the frequency of traffic signal violations at intersections. Both SR, CR and VR are calculated every 10 episodes during both training and testing to provide a consistent assessment of safety performance. We additionally track the total number of collisions (TNC) to measure training-time risk exposure.

In safety-critical scenarios (Scenario~5-6), we evaluate system performance using three key metrics: the success rate, the average vehicle speed, and the mean absolute acceleration. The success rate measures the percentage of completed trials without safety violations or failures. The average vehicle speed reflects overall traffic efficiency, while the mean absolute acceleration quantifies ride comfort and the smoothness of ego vehicle maneuvers.

\begin{table}[t]
\renewcommand{\arraystretch}{1.3}
\centering
\caption{Assessment Results of Different Autonomous Driving Agents in the SUMO Scenarios.}
\label{tab:somo_assessment}
\begin{tabular}{c|cccc}
\hline
\multicolumn{5}{c}{Scenario 3 - Flow 2} \\
\hline
Method & DS & SR & CR & VR \\
\hline
IDM      & 0.83 (0.10) & 0.93 (0.12) & \textbf{0.03 (0.07)} & 0.01 (0.05) \\
SAC      & 0.84 (0.08) & 0.74 (0.13) & 0.26 (0.11) & \textbf{0.00 (0.00)} \\
SAC-Lag  & 0.88 (0.07) & 0.79 (0.11) & 0.21 (0.09) & \textbf{0.00 (0.00)} \\
SMBPO    & 0.84 (0.08) & 0.82 (0.10) & 0.16 (0.08) & 0.02 (0.02) \\
FNI-RL   & 0.94 (0.05) & 0.90 (0.07) & 0.09 (0.07) & 0.01 (0.02) \\
Proposed & \textbf{0.98 (0.02)} & \textbf{0.96 (0.04)} & \textbf{0.03 (0.04)} & 0.01 (0.01) \\
\hline
\multicolumn{5}{c}{Scenario 4} \\
\hline
Method & DS & SR & CR & VR \\
\hline
IDM      & 0.45 (0.17) & 0.46 (0.22) & 0.32 (0.19) & 0.22 (0.19) \\
SAC      & 0.70 (0.12) & 0.72 (0.16) & 0.24 (0.13) & 0.04 (0.12) \\
SAC-Lag  & 0.76 (0.11) & 0.81 (0.14) & 0.17 (0.12) & 0.02 (0.08) \\
SMBPO    & 0.74 (0.09) & 0.79 (0.15) & 0.19 (0.13) & 0.02 (0.05) \\
FNI-RL   & 0.81 (0.10) & 0.88 (0.11) & 0.12 (0.10) & \textbf{0.00 (0.00)} \\
Proposed & \textbf{0.90 (0.06)} & \textbf{0.95 (0.07)} & \textbf{0.05 (0.07)} & \textbf{0.00 (0.00)} \\
\hline
\end{tabular}
\end{table}

\begin{figure*}
    \begin{center}
    \includegraphics[width=0.85\linewidth]{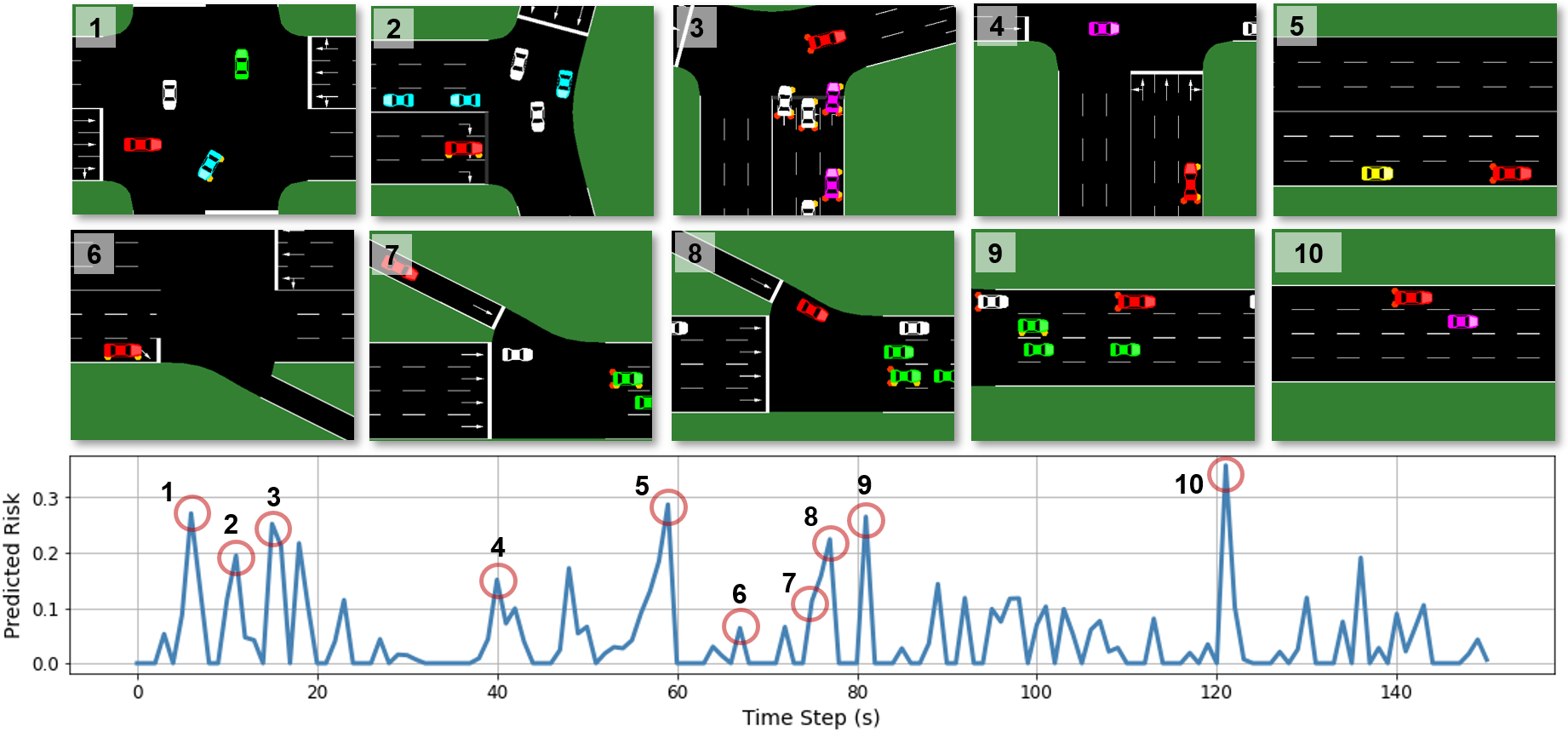}
    \caption{Illustration of real-time risk prediction by the trained world model during a test episode in Scenario~4. Snapshots 1–10 correspond to key traffic interactions, with predicted risk levels visualized over time. High-risk spikes align with complex or uncertain vehicle interactions, demonstrating the model’s capacity to anticipate potential hazards.} 
    \label{P_risk}
 \end{center}                                       
\end{figure*}

\begin{figure} [t!]
\centering
    \subfloat[]{
        \includegraphics[width=0.48\linewidth]{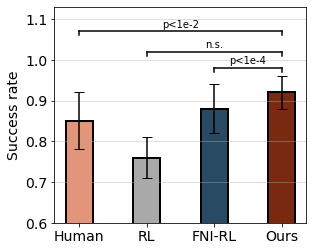}}
    \subfloat[]{
        \includegraphics[width=0.48\linewidth]{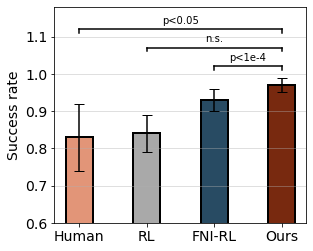}} \\
    \subfloat[]{
        \includegraphics[width=0.47\linewidth]{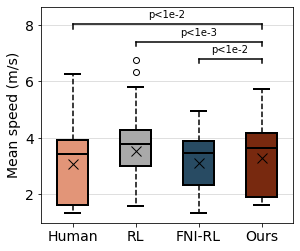}} 
    \subfloat[]{
        \includegraphics[width=0.48\linewidth]{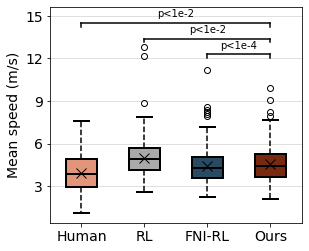}} \\
    \subfloat[]{
        \includegraphics[width=0.48\linewidth]{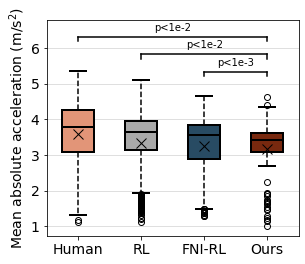}} 
    \subfloat[]{
        \includegraphics[width=0.48\linewidth]{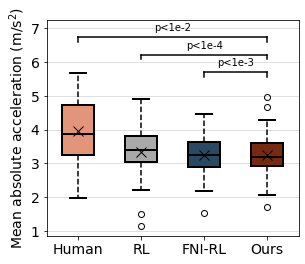}} 
    \caption{Performance evaluation in safety-critical scenarios. Each row compares different approaches on one performance metric: (a)-(b) show the success rate, (c)-(d) the mean speed, and (e)-(f) the mean absolute acceleration. The left column corresponds to the unprotected left turn scenario (Scenario~5), and the right column to the highway emergency scenario (Scenario~6).}
    \label{result_safety_critical} 
\end{figure}

\section{ Results and Discussion}\label{Sec_Results}
\subsection{Evaluation of Regular Traffic Scenarios}
\subsubsection{Training Evaluation}

Fig. \ref{training_env8} illustrates the training performance of different AD algorithms in consecutive signalized intersections (Scenario 3, Flow-1 and Flow-2). Table \ref{tab:test_somo} provides key metric statistics across all scenarios. In Scenario 3, the proposed GTR2L converges faster and achieves the highest success rate with the lowest collision and violation rates under both traffic flows. FNI-RL ranks second with slightly lower performance and training efficiency. SMBPO improves success rate quickly but incurs higher collisions and violations. SAC and SAC-Lag perform worst, especially under high traffic density, showing large fluctuations and high collision risk.

In the long-distance complex Scenario 4 (Fig.~\ref{training_env6}), GTR2L learns rapidly and maintains the highest success rate (~0.94) in later training, significantly outperforming baselines. FNI-RL performs well but with higher collision rates and lower success (~0.85). SMBPO improves success rate but with elevated safety risks. SAC and SAC-Lag perform poorly, struggling to adapt to dynamic environments. GTR2L also achieves the lowest collision, violation rates, and TNC ($2.84 \times 10^2$), demonstrating effective handling of complex interactions.

Overall, GTR2L achieves the best performance across scenarios. In unsignalized intersections (Scenarios 1 and 2), it consistently attains success rates above 0.98, with collision and TNC metrics among the lowest, especially under high-density conditions where it effectively suppresses safety risks. FNI-RL adapts robustly but slightly lags on some metrics. SMBPO and SAC-Lag show moderate results, while SAC’s safety and efficiency remain unsatisfactory in complex scenarios.

\subsubsection{Analysis of Test Results}

Table~\ref{tab:somo_assessment} presents evaluation results in testing scenarios. In the high-density consecutive signalized intersection scenario (Scenario 3 - Flow 2), the proposed method achieves the best performance across all metrics: DS and SR reach 0.98 and 0.96, respectively, with CR and VR both at 0.01. In contrast, the rule-based IDM method, while relatively safe, suffers from lower DS due to limited path planning and efficiency. RL baselines like FNI-RL and SMBPO perform well on some metrics but fall short in overall safety and efficiency. In the long-distance complex navigation scenario (Scenario 4), the proposed method again achieves the highest DS (0.90) and SR (0.95), with CR and VR below 0.01, far surpassing other methods. Other RL baselines show decreased performance, and IDM performs poorly in both safety and efficiency, making it unsuitable for complex environments.

Fig.\ref{P_risk} presents a sequence of key moments from a test episode in Scenario~4, demonstrating how the proposed GTR2L method leverages the trained world model to predict risk in real time. Peaks in the predicted risk curve correspond to complex or uncertain traffic interactions. For instance, significant risk increases are observed when the ego vehicle enters an unsignalized intersection with oncoming traffic (Snapshot 1), navigates a roundabout with potential vehicle cut-ins (Snapshots 2-3), or brakes near a red light while being closely followed (Snapshot 5). Moderate risk arises during highway merging maneuvers (Snapshots 7-8), while low risk is maintained in relatively unoccupied areas, such as a highway ramp (Snapshot 6). Further spikes occur as the vehicle yields on the highway (Snapshot 9) and when adjacent vehicles pass at high speed during cruising (Snapshot 10), highlighting the model’s ability to capture both longitudinal and lateral interaction risks. Overall, by effectively predicting and avoiding potential risks, the proposed GTR2L better adapts to dynamic traffic situations, demonstrating superior decision-making and risk control capabilities.

\begin{figure}[h!]
\centering
    \subfloat[]{
        \includegraphics[width=0.9\linewidth]{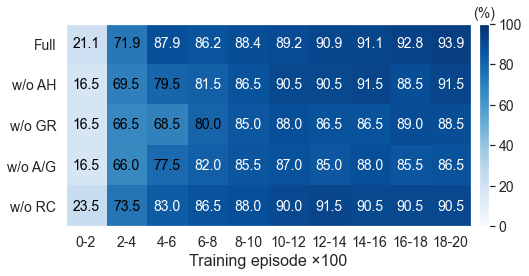}} \\
    \subfloat[]{
        \includegraphics[width=0.9\linewidth]{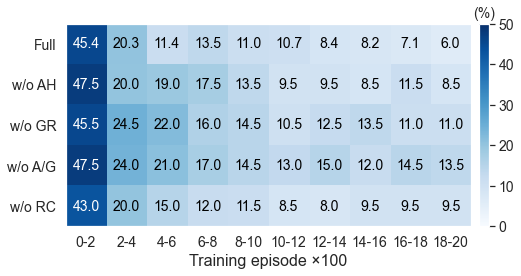}}
    \caption{Training process of the ablation study. (a) Success rate (\%) and (b) collision rate (\%) of different ablation settings under SUMO scenario (d).} 
    \label{fig_ablation}                                
\end{figure}

\subsection{Evaluation of Safety-Critical Scenarios}

We systematically evaluated the performance of different approaches in two representative safety-critical scenarios. In Scenario~5, an unprotected left-turn intersection, Fig.~\ref{result_safety_critical}(a) shows that our method achieves the highest success rate ($p < 1\mathrm{e}{-3}$), indicating significantly improved safety. As shown in Fig.~\ref{result_safety_critical}(e), our approach also achieves the lowest mean absolute acceleration, resulting in smoother driving. Although the RL method attains the highest mean speed in Fig.~\ref{result_safety_critical}(c), its aggressive strategy leads to a much higher collision rate and reduced safety. Human drivers tend to drive at lower speeds but exhibit less smoothness, with higher average acceleration, indicating more conservative and unstable behavior in complex traffic. The FNI-RL method achieves a high success rate but operates at lower speeds, reflecting a conservative strategy that prioritizes safety over efficiency.

In the highway emergency scenario (Scenario~6), Fig.~\ref{result_safety_critical}(b) shows that human drivers have a noticeably lower success rate, highlighting the challenge of handling high-speed emergencies. Our method significantly improves safety in such conditions ($p < 1\mathrm{e}{-4}$), demonstrating strong risk avoidance and emergency response. While FNI-RL performs reasonably, it still yields a lower success rate. As shown in Fig.~\ref{result_safety_critical}(d), human drivers reduce speed out of caution, sacrificing efficiency, while RL pursues higher speeds at the cost of safety. The FNI-RL strategy is more conservative, with lower mean speeds. In contrast, our method maintains both a high success rate and efficient, stable speed, achieving a good balance between safety and efficiency. Fig.~\ref{result_safety_critical}(f) shows that human drivers and RL methods experience greater acceleration fluctuations, leading to less comfortable driving. Though FNI-RL shows improvement, our method ensures the smoothest and most comfortable driving.

\subsection{Ablation Study}

To evaluate the impact of each module on algorithm performance, we conducted ablation experiments in Scenario 4. We compared the full method (Full) with versions where adaptive horizon (w/o AH), game reasoning (w/o GR), both adaptive horizon and game reasoning (w/o A/G), and reachability constraint (w/o RC) were removed. During training, we recorded the success and collision rates at different stages to assess each module’s contribution to safety and efficiency. 

As shown in Fig.~\ref{fig_ablation}, the full method achieves the fastest convergence, while removing AH or GR significantly reduces exploration efficiency. In the later stages, as the policy stabilizes, all ablated variants exhibit varying degrees of degradation, especially w/o A/G. Collision rates further show that removing key modules notably increases collisions, indicating the critical role of RC, AH, and GR in ensuring system safety. For example, in the final training phase (episodes 1800–2000), the collision rate of the full method is 6.0\%. Removing both AH and GR (w/o A/G) increases this by 7.5\%, while removing only GR or RC increases it by 5\% and 3.5\%, respectively. These results highlight the importance of each module in improving the safety and robustness of AD agents.

\section{Conclusion and Outlook}\label{Sec_Conclusion}

This study presents a game-theoretic risk-shaped RL framework for safe AD. By integrating multi-level game-theoretic reasoning, adaptive prediction horizons, and uncertainty-aware reachability constraints, the proposed approach enables robust, risk-aware decision-making in complex traffic environments. Extensive experiments in both urban and safety-critical scenarios show that GTR2L consistently achieves higher success rates, and maintains the lowest collision and violation rates among all compared methods, including human drivers. GTR2L also provides superior efficiency and comfort, with higher mean speeds, smoother driving, and lower acceleration fluctuations, while demonstrating strong adaptability in diverse and challenging scenarios.

Despite these promising results, the proposed framework is currently validated only in simulated environments. Future work will focus on transferring the framework to real-world autonomous vehicles, incorporating more realistic sensor noise and perception uncertainty. Moreover, in the current approach, the policy is fixed after deployment and does not adapt to new situations. This limitation highlights the importance of continual learning and online adaptation mechanisms, which are essential for enabling intelligent agents to respond effectively to long-term environmental changes and unforeseen scenarios.

\bibliographystyle{IEEEtran}

\small\bibliography{Bibliography}
\end{document}